\documentclass[10pt,twocolumn,letterpaper]{article}

\usepackage[pagenumbers]{cvpr} 
\usepackage{multirow}
\usepackage[T1]{fontenc}
\usepackage{caption}
\usepackage{placeins}
\usepackage{xcolor}
\usepackage{cuted} 

\usepackage{currfile} 

\usepackage{caption} 
\definecolor{cvprblue}{rgb}{0.21,0.49,0.74}
\usepackage[pagebackref,breaklinks,colorlinks,allcolors=cvprblue]{hyperref}

\def\confName{CVPR}
\def\confYear{2027}

\title{Beyond Sample Hardness: Adaptive Data Selection through Learning Dynamics}

\author{
Hai Gan$^{\dagger,1}$ \qquad Suorong Yang$^{\dagger,*,1,2}$ \qquad Guang Li$^{3}$\\
Baile Xu$^{1}$ \qquad Furao Shen$^{1}$\\[0.8em]
$^{1}$Nanjing University, Nanjing, China\\
$^{2}$National University of Singapore, Singapore\\
$^{3}$Hokkaido University, Sapporo, Japan\\[0.4em]
{\tt\small \{haigan,sryang\}@smail.nju.edu.cn}\\
{\tt\small \{xubaile,frshen\}@nju.edu.cn}\\
{\tt\small guang@lmd.ist.hokudai.ac.jp}\\[0.4em]
}

\begin{document}
\maketitle
\begin{abstract}
Data selection has emerged as an effective approach to reducing the model training costs. However, most existing methods often rely heavily on sample hardness, which may overlook differences in sample learning dynamics and lead to suboptimal sample selection. In this paper, we propose \textbf{LeDAS}, an end-to-end \textbf{Le}arning \textbf{D}ynamics-\textbf{A}ware \textbf{S}election framework for dynamic data selection, which jointly characterizes sample importance through hardness, learning progress, and deterioration, enabling adaptive sample prioritization throughout training with marginal additional computational overhead. Specifically, we track per-sample loss dynamics using exponential moving averages (EMA) to obtain three complementary signals that capture current learning difficulty, ongoing learning improvements, and potential degradation. To accommodate evolving training dynamics, we introduce Adaptive Metric Fusion, a contextual bandit-based mechanism that dynamically adjusts the relative importance of these signals based on training feedback. Furthermore, a bounded staleness compensation mechanism accounts for historical selection patterns by increasing the selection priority of samples that have not been selected for extended periods. Extensive experiments across diverse vision and language tasks demonstrate the effectiveness and generality of LeDAS. Compared with state-of-the-art baselines, LeDAS consistently achieves superior performance, matching or surpassing full-dataset training at a 50\% selection ratio on ImageNet-1K and in LLM supervised fine-tuning. Moreover, its modular and task-agnostic design enables the integration of complementary selection signals, demonstrating strong robustness to noisy and corrupted datasets.
\end{abstract}

\section{Introduction}
\label{sec:intro}
\begin{figure}[!t]
    \centering
    \includegraphics[width=\linewidth]
    {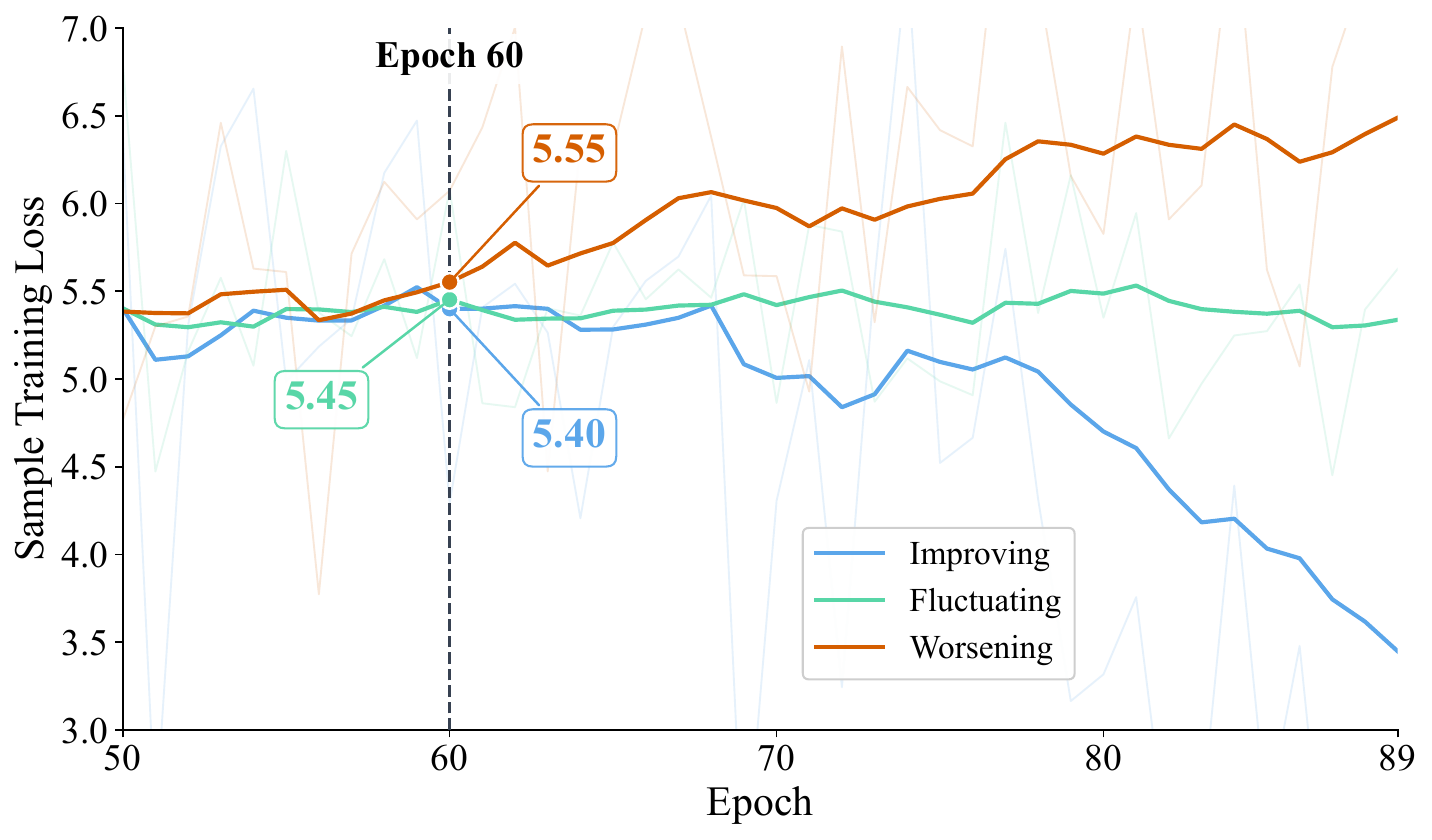}
    \caption{Illustration of distinct learning trajectories on Tiny-ImageNet. Samples with similar EMA-smoothed losses at epoch 60 exhibit diverse subsequent learning behaviors, including improving, fluctuating, or worsening trends. Faint and solid curves represent instantaneous and EMA-smoothed losses, respectively.}
    \label{fig:learning_dynamics}
    \vspace{-5pt}
\end{figure}
Recent advances in deep learning have been driven largely by the growing scale of models and datasets~\cite{transformer, imagenet, llama, gpt4}. While large-scale datasets provide diverse training examples, they also incur substantial computational costs and often contain redundant, noisy, or less informative samples. These challenges have motivated data-efficient learning, which seeks to reduce training costs while preserving model performance. Among existing approaches, data selection methods~\cite{glister, moderate, infobatch, data_agent, dual} identify informative subsets to reduce training costs while preserving or improving model performance. These methods can be broadly categorized into static selection~\cite{data_diet, rl_select, moderate}, which identifies a fixed subset before training, and dynamic selection~\cite{infobatch, ucb, data_agent, orderdp}, which adaptively updates the selected subset to accommodate evolving model states.

A central challenge in dynamic selection is determining which samples deserve further learning at each training stage. Existing methods often estimate sample importance using training loss or prediction uncertainty~\cite{infobatch, data_agent, bls, seta, rho_loss}, assuming that harder samples provide stronger learning signals. However, sample hardness does not necessarily reflect the potential benefit of further training, as samples with similar loss values may exhibit substantially different learning trajectories. As illustrated in Fig.~\ref{fig:learning_dynamics}, samples with comparable losses smoothed by an exponential moving average (EMA) can follow distinct trajectories, including sustained improvement, persistent fluctuations, or increasing loss. Despite these differences, hardness-based methods may assign similar priorities to such samples, potentially wasting training resources. This motivates moving beyond sample hardness to characterize learning dynamics and prioritize samples that contribute more effectively to model improvement.

To this end, we propose \textbf{Le}arning \textbf{D}ynamics-\textbf{A}ware \textbf{S}election (\textbf{LeDAS}), an end-to-end adaptive data selection framework that prioritizes samples according to their evolving learning dynamics. Specifically, LeDAS characterizes sample-level learning behavior through three complementary metrics derived from EMA-smoothed loss statistics: hardness (current learning difficulty), progress (recent loss reduction), and deterioration (recent loss increase). To adapt their relative importance throughout training, we introduce Adaptive Metric Fusion, a contextual bandit-based mechanism that dynamically adjusts their weights. A lightweight MLP policy generates Dirichlet-distributed fusion weights conditioned on globally aggregated feature representations from the target model and is optimized using REINFORCE~\cite{reinforce}, with rewards derived from the relative reduction in full-dataset training loss. Furthermore, bounded Staleness Compensation adjusts sample priorities according to historical selection patterns, encouraging the reconsideration of samples that have remained unselected for extended periods.

Extensive experiments across diverse datasets, architectures, and learning tasks demonstrate the effectiveness and broad applicability of LeDAS. On image classification, LeDAS consistently outperforms existing data selection methods and matches or surpasses full-dataset training on ImageNet-1K using only 50\% of the data. Its generalization capabilities are further evaluated on challenging benchmarks, including ImageNet-A/R/Hard/O. Beyond image classification, LeDAS extends effectively to object detection, semantic segmentation, and LLM supervised fine-tuning, achieving performance comparable to or better than full-dataset fine-tuning across LLM benchmarks. For example, with only 50\% of the training data, LeDAS improves the Pass@1 of Qwen3.5-9B from 82.50\% to 85.00\% on AMC23 and from 60.09\% to 62.63\% on GPQA-Diamond. Moreover, its modular design supports the integration of semantic consistency signals and improves robustness to noisy and corrupted datasets.

Our main contributions are summarized as follows: \textbf{1)} We investigate the limitations of hardness-based data selection and propose a learning dynamics-aware characterization that jointly considers hardness, progress, and deterioration, enabling asymmetric modeling of improving and deteriorating loss trends. \textbf{2)} We introduce Adaptive Metric Fusion, a contextual bandit-based mechanism that dynamically balances learning signals according to training feedback. Together with bounded Staleness Compensation, LeDAS adapts sample prioritization to evolving learning dynamics and historical selection patterns. \textbf{3)} Extensive experiments across diverse datasets, architectures, and tasks demonstrate the effectiveness and generality of LeDAS, achieving substantial training acceleration in vision tasks while maintaining or improving model performance and extending effectively to LLM supervised fine-tuning.

\section{Related Work}
\label{sec:related_work}

Data selection can be broadly classified into: static data selection~\cite{forgetting, moderate, data_diet, submodular} and dynamic data selection~\cite{infobatch, data_agent, rho_loss, ddsa, orderdp, bls, dyn-unc}.
Static methods select a fixed subset before training, whereas dynamic methods adaptively update the selected subset throughout training.

\noindent \textbf{Static Data Selection.}
Existing methods can be broadly categorized into three groups: importance-criteria-based, geometry-based, and optimization-based approaches. Specifically, importance-criteria-based methods rank samples according to predefined importance measures. For example, EL2N and GraNd~\cite{data_diet} estimate sample importance using the $\ell_2$ norms of prediction errors and gradient norms, respectively, while Forgetting~\cite{forgetting} identifies samples based on misclassification frequency. Other methods estimate sample importance through memorization effects~\cite{score-based-3} or sample removal effects~\cite{moso}. In contrast, geometry-based methods select representative samples to preserve the distributional characteristics or coverage of the original dataset. Herding~\cite{herding} selects samples close to class centers, while Moderate~\cite{moderate} prioritizes samples with scores close to the median. CCS~\cite{ccs} extends the classical set cover problem to distribution coverage, while other works~\cite{weighted_k-center} employ k-center selection or uncertainty-aware sampling to improve data coverage. Finally, optimization-based methods formulate data selection as an optimization problem, aiming to identify subsets that preserve important training properties~\cite{yang2024clip}, which include gradient matching~\cite{coreset, opt-based-2}, influence estimation~\cite{dataset_pruning}, bi-level optimization~\cite{glister}, and submodular optimization~\cite{opt-based-4, opt-based-1}. Despite their effectiveness, static selection methods retain a fixed subset throughout training and thus cannot adapt to evolving sample-level learning dynamics.

\noindent \textbf{Dynamic Data Selection.}
Dynamic data selection updates the selected subset during training to adapt to evolving model states. For example, the work~\cite{ucb} employs upper confidence bound (UCB) and $\epsilon$-greedy strategies to guide sample selection. The method in~\cite{dyn-unc} measures prediction variations throughout training to capture both prediction uncertainty and training dynamics, thereby identifying informative samples. RHO-LOSS~\cite{rho_loss} proposes a dynamic data selection method that estimates sample importance using the difference between the current training loss and the irreducible loss predicted by an auxiliary model, thereby prioritizing learnable and task-relevant samples while reducing the selection of noisy and redundant data. SeTa~\cite{seta} adopts a curriculum learning strategy that selects high-value samples in an easy-to-hard progression, enabling lossless acceleration in large-scale training. InfoBatch~\cite{infobatch} introduces an unbiased dynamic data pruning framework to randomly remove less informative samples based on the loss distribution. Recently, BLS\cite{bls} employs an exponential moving average of batch losses to approximate individual sample importance, while OrderDP~\cite{orderdp} establishes unbiased gradient estimation uniformly samples candidates and selects the top-$q$ samples based on maintained scores. Despite these advances, existing methods rely on different selection criteria, such as uncertainty, prediction variations, curriculum strategies, or loss-based importance estimates, leaving room for more comprehensive characterization of sample-level learning dynamics.

\section{The Proposed Method}

\begin{figure*}[!t]
    \centering
    \includegraphics[width=.85\linewidth]{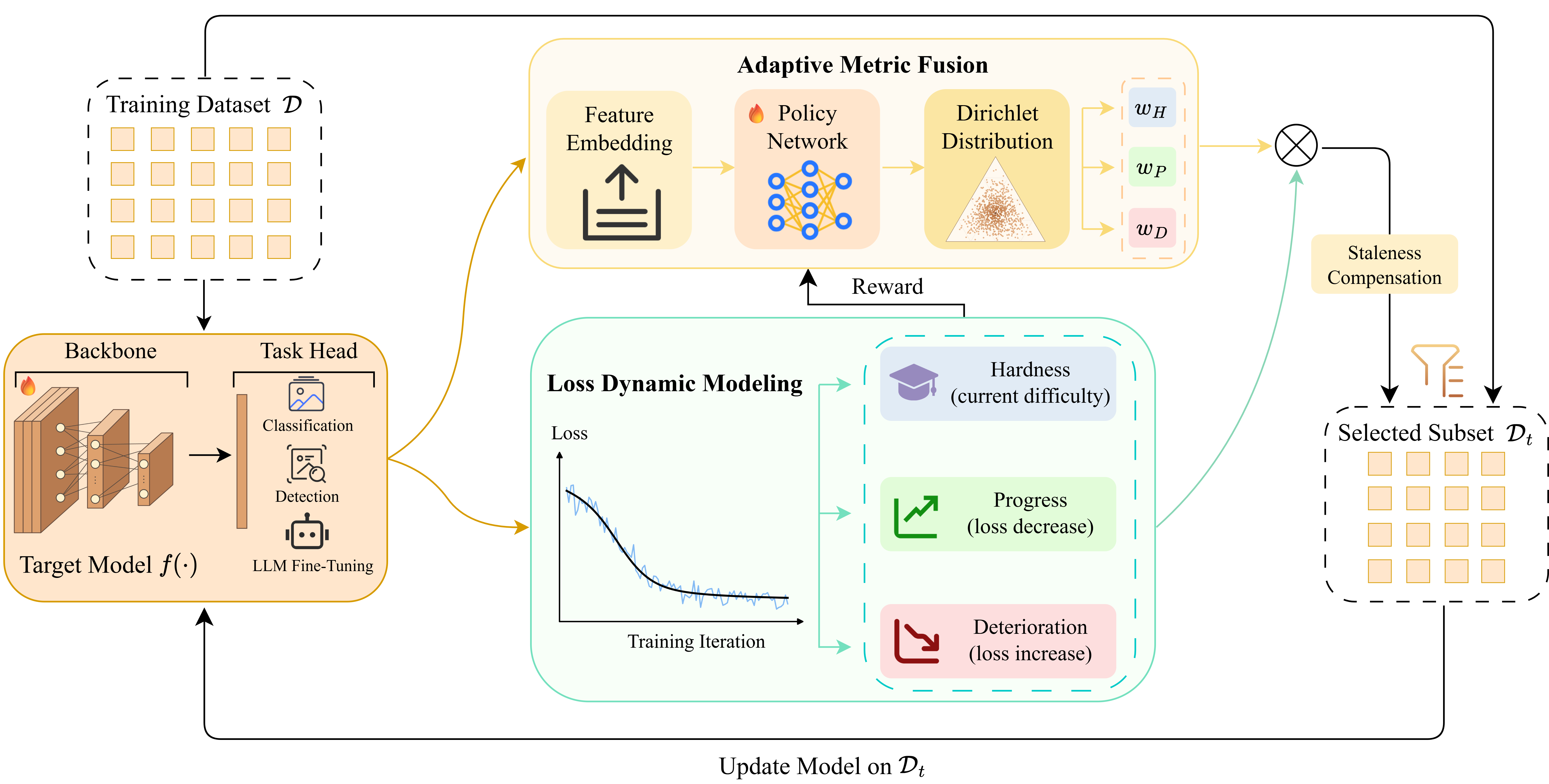}
    \caption{Overview of the proposed LeDAS framework. At each training epoch, LeDAS characterizes sample-level learning dynamics using three metrics: hardness, progress, and deterioration. It then adaptively fuses these signals via a contextual bandit guided by target model features. Staleness Compensation further adjusts sample selection priorities based on historical selection patterns. The selected subset is finally used to update the target model, thereby establishing a closed-loop data selection framework.}
    \label{fig:overview}
    \vspace{-5pt}
\end{figure*}

As illustrated in \cref{fig:overview}, we propose LeDAS, a Learning Dynamics-Aware data Selection framework that characterizes sample learning behavior through three loss-derived signals: hardness, progress, and deterioration. LeDAS employs Adaptive Metric Fusion to dynamically balance these signals through a feature-conditioned contextual bandit, and further incorporates Staleness Compensation to account for historical selection patterns. These components jointly enable adaptive sample prioritization throughout training.

\subsection{Preliminaries and Problem Formulation}
\label{sec:preliminaries}

Let $\mathcal{D}=\{(x_i,y_i)\}_{i=1}^{N}$ be a training dataset, and let $f(\cdot;\theta_t)$ denote the model at selection round $t$. For a sample $(x,y)\in\mathcal{D}$, its loss is
\begin{equation}
    \ell_t(x,y)=\ell\big(f(x;\theta_t),y\big).
    \label{eq:sample_loss}
\end{equation}


Given a selection ratio $\rho\in(0,1]$, a dynamic selection policy retains $K=\lfloor\rho N\rfloor$ samples in each round, forming a subset $\mathcal{D}_t\subseteq\mathcal{D}$ for the next model update.
Let $\theta_{\mathrm{full}}$ and $\theta_{\mathrm{sel}}$ denote the final model parameters obtained by training on the full dataset and dynamically selected subsets, respectively. For the underlying data distribution $\mathcal{P}$, the expected risk is
\begin{equation}
    \mathcal{R}(\theta)
    =\mathbb{E}_{(x,y)\sim\mathcal{P}}
      \big[\ell(f(x;\theta),y)\big].
    \label{eq:expected_risk}
\end{equation}

Our objective is to minimize the risk discrepancy between compressed and full-data training:
\begin{equation}
    \min_{\pi}\;
    \mathbb{E}_{\pi}\!
    \left[
        \mathcal{R}(\theta_{\mathrm{sel}})
        -\mathcal{R}(\theta_{\mathrm{full}})
    \right],
    \label{eq:compression_objective}
\end{equation}
where $\pi$ denotes the dynamic selection policy and $\theta_{\mathrm{sel}}$ depends on $\pi$ and $\rho$.

\subsection{Learning Dynamics Modeling}
\label{sec:learning_dynamics}

For each sample, we track its loss change
$\Delta\ell_t=\ell_t-\ell_{t-1}$ and smooth both the loss
and its change using EMA:
\begin{align}
    \bar{\ell}_t
    &= \beta_\ell\bar{\ell}_{t-1}
       +(1-\beta_\ell)\ell_t, \notag\\
    \bar{\Delta}_t
    &= \beta_\Delta\bar{\Delta}_{t-1}
       +(1-\beta_\Delta)\Delta\ell_t.
    \label{eq:ema_dynamics}
\end{align}
Based on these statistics, we characterize sample learning
dynamics through three metrics:
\begin{equation}
    H_t = \bar{\ell}_t, \quad
    P_t = -\bar{\Delta}_t, \quad
    D_t = \max\!\big(\bar{\Delta}_t,0\big).
    \label{eq:hpd_metrics}
\end{equation}

Specifically, hardness $H_t$ characterizes the current learning difficulty, progress $P_t$ captures ongoing learning improvements, and deterioration $D_t$ indicates potential learning degradation that may require further attention. Unlike hardness, which reflects the current loss magnitude, progress and deterioration characterize directional changes in sample learning behavior. Although $P_t$ and $D_t$ are derived from the same loss-change signal, their separate formulations enable asymmetric weighting of improving and deteriorating trends, as detailed in Appendix~\ref{app:asymmetric_dynamics}.


\subsection{Bandit Policy Gradient Analysis}
\label{sec:bandit_policy}
 
\paragraph{Adaptive Metric Fusion.}
To adaptively learn the relative importance of the three signals throughout training, we introduce a learnable contextual bandit. Specifically, we extract penultimate-layer feature representations from the target model, corresponding to the core representations immediately preceding its task-specific prediction head, and aggregate them across the training dataset to construct a global context:
\begin{equation}
    \mathbf{c}_t =
    \frac{1}{N}\sum_{i=1}^{N}
    f^{\mathrm{feat}}(x_i;\theta_t),
    \quad
    \mathbf{c}_t\in\mathbb{R}^{d},
    \label{eq:bandit_context}
\end{equation}
where $f^{\mathrm{feat}}(\cdot;\theta_t)$ denotes the
feature extraction function of the target model at
training step $t$, $N$ is the number of training samples,
and $d$ is the feature dimension.

We parameterize the policy network $g(\cdot;\phi)$ with a lightweight MLP to map the global context into Dirichlet concentration parameters. The resulting parameters and stochastic fusion weights are obtained as follows:
\begin{equation}
    \boldsymbol{\alpha}_t
    =\operatorname{Softplus}\left(
    g(\mathbf{c}_t;\phi)\right),
    \quad
    \mathbf{w}_t\sim
    \operatorname{Dirichlet}(\boldsymbol{\alpha}_t),
    \label{eq:dirichlet_policy}
\end{equation}
where $\mathbf{w}_t=(w_t^P,w_t^H,w_t^D)$ denotes the sampled fusion weights. While the Softplus transformation guarantees positive concentration parameters, the Dirichlet distribution restricts the fusion weights to the probability simplex. This not only enforces non-negativity and normalization but also facilitates stochastic exploration across different metric combinations. Before computing the base selection score, we independently normalize each learning dynamics metric using Min-Max normalization to obtain comparable scales. The resulting normalized metrics are then combined using the sampled fusion weights to compute the base selection score:
\begin{equation}
    s_t^{\mathrm{base}}
    =w_t^P\widetilde{P}_t
    +w_t^H\widetilde{H}_t
    +w_t^D\widetilde{D}_t.
    \label{eq:adaptive_selection_score}
\end{equation}

\paragraph{Training Feedback and Policy Update.}
We define the reward as the relative reduction in full-dataset training loss. Specifically, the average training loss at round $t$ is given by $\mathcal{L}_t=\frac{1}{N}\sum_{i=1}^{N}\ell_t(x_i,y_i)$. After updating the target model on the selected subset and obtaining $\theta_{t+1}$, we evaluate the full-dataset loss again and compute the reward as
\begin{equation}
    r_t=
    \frac{\mathcal{L}_t-\mathcal{L}_{t+1}}
         {|\mathcal{L}_t|+\epsilon},
    \label{eq:bandit_reward}
\end{equation}
where $\epsilon>0$ is introduced for numerical stability. The full-dataset loss is evaluated without gradient computation, providing feedback on the effectiveness of the fusion weights sampled at round $t$.

We optimize the Dirichlet policy using REINFORCE~\cite{reinforce} with entropy regularization. The policy objective is formulated as
\begin{equation}
    J_t(\phi)=
    \mathbb{E}_{\mathbf{w}_t\sim\pi_\phi(\cdot\mid\mathbf{c}_t)}[r_t]
    +\lambda_{\mathrm{ent}}\,
    \mathcal{H}\!\left(\pi_\phi(\cdot\mid\mathbf{c}_t)\right),
    \label{eq:bandit_objective}
\end{equation}
where $\pi_\phi(\cdot\mid\mathbf{c}_t)$ denotes the conditional Dirichlet policy, $\mathcal{H}(\cdot)$ represents its entropy, and $\lambda_{\mathrm{ent}}$ controls the strength of entropy regularization to encourage exploration. Following the likelihood-ratio trick, the policy gradient is estimated as
\begin{equation}
\begin{aligned}
    \widehat{\nabla_\phi J_t}
    ={}&(r_t-b_t)\nabla_\phi
       \log\pi_\phi(\mathbf{w}_t\mid\mathbf{c}_t)\\
      &+\lambda_{\mathrm{ent}}\nabla_\phi
       \mathcal{H}\!\left(\pi_\phi(\cdot\mid\mathbf{c}_t)\right),
\end{aligned}
\label{eq:reinforce_gradient}
\end{equation}
where $b_t$ denotes an EMA baseline of previous rewards, updated as $b_{t+1}=\mu_b b_t+(1-\mu_b)r_t$, with $\mu_b$ controlling the smoothing rate. The policy parameters $\phi$ are optimized using Adam~\cite{adam}. Under standard likelihood-ratio assumptions, this estimator provides an unbiased gradient estimate of the regularized policy objective, as established in Appendix~\ref{app:reinforce_proof}.

\begin{table*}[t]
    \centering
    \caption{Comparison of classification accuracy (\%) against state-of-the-art baselines. All methods use ResNet-18 on CIFAR-10/100 and ResNet-50 on Tiny-ImageNet. Some results are unavailable due to the absence of publicly released code or sufficient implementation details for reproduction. Random* denotes random sample selection at each epoch.\label{tab:cifar_tiny}}
	\resizebox{0.62\textwidth}{!}{
    \begin{tabular}{c|ccc|ccc|ccc}
    \toprule[1.5pt]
    Dataset &  \multicolumn{3}{c|}{ CIFAR-10}& \multicolumn{3}{c|}{ CIFAR-100} & \multicolumn{3}{c}{ Tiny-ImageNet} \\ \hline
    Full Dataset &\multicolumn{3}{c|}{95.6}&\multicolumn{3}{c|}{78.2} &\multicolumn{3}{c}{45.0} \\ \hline
    Selection Ratio (\%)& 30& 50&70& 30& 50&70& 30& 50&70 \\ \hline
    
    Random &90.2&92.3&93.9&69.7&72.1&73.8&29.8&37.2&42.2 \\ 
    Herding~\cite{herding} &80.1 &88.0&92.2 &69.6&71.8&73.1&29.4&31.6&39.8 \\
    EL2N~\cite{data_diet} &91.6&95.0&95.2 &69.5&72.1&77.2 &26.6&37.1&44.0 \\
    GraNd~\cite{data_diet} &91.2&94.6&95.3 &68.8&71.4&74.6 &29.7&36.3&43.2 \\
    Glister~\cite{glister} &90.9&94.0&95.2 &70.4&73.2&76.6 &30.1&39.5&43.9 \\
    Forgetting~\cite{forgetting} &91.7&94.1&94.7 &69.9&73.1&75.3 &28.7&33.0&41.2 \\
    Moderate-DS~\cite{moderate} &91.5&94.1&95.2&70.2&73.4&77.3 &30.6&38.2&42.8 \\
    Self-sup. prototypes~\cite{beyond} &91.0&94.0&95.2&70.0&71.7&76.8 &27.7&37.9&43.4 \\
    RL-Selector~\cite{rl_select} &91.8&-&95.4 &71.1&-&77.6 &31.1&-&44.5 \\
    MoSo~\cite{moso}  &91.1&94.2&95.3 &70.9&73.6&77.5 &31.2&38.5&43.4 \\
    DP~\cite{dataset_pruning}&90.8&93.8&94.9 &-&73.1&77.2 &-&-&- \\
    Random* &93.0&94.5&94.8 &74.4&75.3&77.3 &41.5&42.8&43.1 \\
    UCB~\cite{ucb} &93.9&94.7&95.3 &-&75.3&77.3 &-&-&-    \\
    $\epsilon$-Greedy~\cite{ucb} &94.1&94.9&95.2 &-&74.8&76.4&-&-&- \\
    CLIP-Sel~\cite{yang2024clip} &91.9&94.5&95.0 &70.8&73.7&77.0 &31.7&40.0&46.0 \\ 
    DUAL~\cite{dual} &91.8&95.4&95.3 &66.4&74.6&77.4 &-&-&- \\ 
    OrderDP~\cite{orderdp} &95.0&95.3&95.6 &76.7&77.9&78.2 &-&-&- \\ 
    InfoBatch~\cite{infobatch} &94.7&95.1&95.6 &76.5&78.1&78.2 &42.2&43.2&43.8 \\
    Data-Agent~\cite{data_agent} &95.0&95.3&\textbf{96.0} &77.6&78.9&79.5 &44.9&47.0&49.4 \\ 
    \hline
    LeDAS&\textbf{95.3}&\textbf{95.6}&95.7&\textbf{78.6}&\textbf{79.4}&\textbf{79.8}&\textbf{45.9}&\textbf{49.1}&\textbf{50.2} \\
    \bottomrule[1.5pt]
    \end{tabular}}
\end{table*}

\subsection{Staleness Compensation}
\label{sec:staleness}

To account for historical selection patterns, we introduce a staleness-aware compensation mechanism. For a fixed sample, let $t'$ denote the round in which it was most recently selected. Its staleness age at round $t$ is
\begin{equation}
    \nu_t=t-t',
    \label{eq:staleness_age}
\end{equation}
Subsequently, we apply a saturating multiplicative bonus:
\begin{equation}
    B_t=1+\lambda\left(1-\exp\left(-\frac{\nu_t}{\tau}\right)\right),
    \label{eq:staleness_bonus}
\end{equation}
and obtain the final score by
\begin{equation}
    s_t^{\mathrm{final}}=s_t^{\mathrm{base}}\,B_t.
    \label{eq:staleness_final_score}
\end{equation}

The training subset is then constructed by selecting the samples with the highest final scores. For $\lambda>0$ and $\tau>0$, the compensation is nondecreasing and bounded by:
\begin{equation}
\begin{aligned}
    1&\leq B_t<1+\lambda,\\
    \frac{\partial B_t}{\partial\nu_t}
    &=\frac{\lambda}{\tau}
      \exp\left(-\frac{\nu_t}{\tau}\right)\geq 0.
\end{aligned}
\label{eq:staleness_properties}
\end{equation}
Thus, the staleness compensation factor increases monotonically
with staleness age while remaining bounded between $1$ and
$1+\lambda$, allowing historical selection patterns to influence
sample priorities without excessively amplifying base scores.
The proofs of monotonicity and boundedness are deferred to the
Appendix~\ref{app:staleness_proof}.

\begin{table*}[]
\centering
\caption{Training efficiency on ImageNet-1K with ResNet-50 at a 60\% selection ratio, using 8 A100 GPUs. We report wall-clock time (h) and total GPU time (GPU hours). Glister~\cite{glister} and CG-Score~\cite{cgscore} are omitted due to their high computational and memory costs. Some results are taken from InfoBatch~\cite{infobatch}. \label{tab:efficiency}}
\resizebox{\textwidth}{!}{
    \begin{tabular}{lcccccccccccc}
    \toprule
    Method
    & Herding
    & EL2N
    & GraNd
    & Forgetting
    & RL-Selector
    & SSP
    & Moderate
    & CLIP-Sel
    & UCB
    & Infobatch
    & LeDAS
    & Full Dataset \\
    \midrule
    Acc. (\%)
    & 71.1 & 72.3 & 71.0 & 72.5 & 73.4 & 70.0
    & 73.1 & 73.2 & 75.8 & 76.5
    & \textbf{76.7} & 76.4 \\
    
    Time (h)
    & 10.5 & 10.5 & 10.5 & 10.5 & 10.5 & 10.5
    & 10.5 & 10.5 & 10.5 & 10.5
    & 10.5 & 17.5 \\
    
    Overhead (h)
    & $>17.5$ & $>17.5$ & $>17.5$ & $>17.5$
    & $>17.5$ & $>24.0$ & $>17.5$ & $>1.6$
    & 0.03 & 0.0028 & 0.114 & 0.0 \\
    
    Overall ($n^*h$)
    & $>224.0$ & $>224.0$ & $>224.0$ & $>224.0$
    & $>224.0$ & $>276.0$ & $>224.0$ & $>96.8$
    & 84.2 & 84.0 & 84.9 & 140.0 \\
    \bottomrule
    \end{tabular}
    }
\end{table*}


\begin{table*}
    \centering
    \begin{minipage}[]{0.54\textwidth}
        \caption{Generalization across different model architectures on ImageNet-1K, including ResNet-50, Swin-Base, and ViT-Large. All experiments are conducted on a server equipped with 4 A100 GPUs. $S_r$ denotes the data selection ratio.}
            \label{tab:imagenet1k}
            \resizebox{.99\textwidth}{!}{%
            \begin{tabular}{c|ccccccc|c}
                \toprule[0.8pt]
                $S_r(\%)$ & 20 & 30 & 40 & 50 & 60 & 70 & 80 & Full Dataset \\
                \midrule[0.4pt]
        
                ResNet-50
                & 74.6 & 75.4 & 76.2 & 76.4 & 76.7 & 77.3 & \textbf{77.5} & 76.4 \\
        
                Swin-Base
                & 84.0 & 84.1 & 84.0 & 84.1 & 84.1 & \textbf{84.2} & \textbf{84.2} & 84.2 \\
        
                ViT-Large
                & 84.9 & 84.9 & 85.0 & 85.0 & 85.1 & 85.3 & \textbf{85.4} & 84.6 \\
        
                \bottomrule[0.8pt]
            \end{tabular}}
        \end{minipage}
        \hspace{12pt}
         \begin{minipage}[]{0.35\textwidth}
                 \centering
            \caption{Object detection performance on MS-COCO using YOLOv8 and RT-DETR under different selection ratios. Results are reported as AP$_{50}$/AP (\%).}
            \label{tab:coco}
        
            \footnotesize
            \setlength{\tabcolsep}{3pt}
            \renewcommand{\arraystretch}{1.15}
            \resizebox{.99\textwidth}{!}{%
                \begin{tabular}{ll|ccc|c}
            \toprule[0.8pt]
            Model & Method & 70\% & 80\% & 90\% & Full Dataset \\
            \midrule[0.4pt]
    
            \multirow{2}{*}{YOLOv8}
            & Random
            & 38.4/26.1 & 39.5/27.1 & 40.1/27.4
            & \multirow{2}{*}{40.4/27.6} \\
            & LeDAS
            & \textbf{38.7/26.4}
            & \textbf{39.7/27.2}
            & \textbf{40.8/28.0}
            & \\
    
            \midrule[0.4pt]
    
            \multirow{2}{*}{RT-DETR}
            & Random
            & 55.3/40.9 & 56.3/41.1 & 57.1/42.2
            & \multirow{2}{*}{57.5/42.4} \\
            & LeDAS
            & \textbf{55.8/41.2}
            & \textbf{56.7/41.5}
            & \textbf{57.5/42.7}
            & \\
    
            \bottomrule[0.8pt]
        \end{tabular}%
        }
         \end{minipage}
    
\end{table*}







\section{Experiment}

\subsection{Experimental Setup}
\label{sec:experimental_setup}

\textbf{Datasets and Tasks.}
We evaluate LeDAS across four representative tasks: image
classification, object detection, semantic segmentation, and
large language model (LLM) supervised fine-tuning.
For image classification, we conduct experiments on
CIFAR-10/100~\cite{cifar100}, Tiny-ImageNet~\cite{tiny},
and ImageNet-1K~\cite{imagenet}. To evaluate robustness and
generalization under challenging distribution shifts, we
further consider ImageNet-A/O~\cite{imagenet-a},
ImageNet-R~\cite{imagenet-r}, and
ImageNet-Hard~\cite{imagenet-hard}. We also evaluate LeDAS
on the long-tailed recognition benchmarks ImageNet-LT and
Places-LT~\cite{lt_datasets}. For object detection and
semantic segmentation, we use MS-COCO~\cite{coco} and
ADE20K~\cite{ade20k}, respectively. For LLM supervised fine-tuning, we use Nemotron-Math~\cite{nemotron_math}
as the training dataset and evaluate model performance on
five benchmarks: AIME 2024~\cite{aime2024}, AMC 2023~\cite{amc23},
Olympiad~\cite{olympiad}, GPQA-Diamond~\cite{gpqa}, and
MMLU-Pro~\cite{mmlu_pro}.

\noindent\textbf{Model Architectures.}
We evaluate LeDAS across diverse model architectures to
assess its applicability to different network designs.
For image classification, we consider both convolutional
networks and Transformer-based architectures, including
ResNet-18/50~\cite{resnet}, ViT-B/L~\cite{vit}, and
Swin Transformer~\cite{swin-t}. For object detection, we
employ YOLOv8~\cite{yolov8} and RT-DETR~\cite{detr},
while UPerNet~\cite{upernet} and PSPNet~\cite{pspnet}
are used for semantic segmentation. For LLM supervised fine-tuning, we adopt Qwen3-4B~\cite{qwen3_4b} and
Qwen3.5-9B~\cite{qwen3_5_9b} to assess the effectiveness
of LeDAS across different model scales.

\noindent\textbf{Comparison with State-of-the-art Baselines.}
To evaluate the effectiveness of LeDAS, we compare it with
17 representative data selection methods, including
1) Herding~\cite{herding},
2) EL2N~\cite{data_diet},
3) GraNd~\cite{data_diet},
4) Glister~\cite{glister},
5) Forgetting~\cite{forgetting},
6) Moderate-DS~\cite{moderate},
7) Self-sup. Prototypes~\cite{beyond},
8) RL-Selector~\cite{rl_select},
9) MoSo~\cite{moso},
10) DP~\cite{dataset_pruning},
11) UCB~\cite{ucb},
12) $\epsilon$-Greedy~\cite{ucb},
13) CLIP-Sel~\cite{yang2024clip},
14) DUAL~\cite{dual},
15) OrderDP~\cite{orderdp},
16) InfoBatch~\cite{infobatch}, and
17) Data-Agent~\cite{data_agent}.

\subsection{Performance Comparison}

\noindent\textbf{Results on CIFAR-10/100 and Tiny-ImageNet.}
As shown in \cref{tab:cifar_tiny}, LeDAS achieves superior performance over existing data selection methods in most settings across the three datasets. At a 50\% selection ratio, LeDAS obtains accuracies of 95.6\%, 79.4\%, and 49.1\% on CIFAR-10, CIFAR-100, and Tiny-ImageNet, respectively, surpassing the strongest competing baselines by 0.2, 0.5, and 2.1 percentage points. Notably, even with only 30\% of the training samples, LeDAS achieves 78.6\% on CIFAR-100 and 45.9\% on Tiny-ImageNet, exceeding their full-dataset baselines. Increasing the selection ratio to 70\% further enables LeDAS to surpass full-dataset training across all three datasets, achieving 95.7\%, 79.8\%, and 50.2\% accuracy, respectively. These results demonstrate the effectiveness of learning dynamics-aware selection in preserving and improving model performance under substantially reduced training data budgets.

\noindent\textbf{Results on ImageNet-1K.}
We evaluate the effectiveness and efficiency
of LeDAS on large-scale ImageNet-1K.
As shown in \cref{tab:efficiency},
with a 60\% selection ratio, LeDAS achieves 76.7\%
Top-1 accuracy, outperforming InfoBatch and full-dataset
training by 0.2 and 0.3 percentage points, respectively.
Meanwhile, LeDAS reduces the total GPU time from 140.0
to 84.9 GPU hours, achieving a 39.4\% reduction in
computational cost. With a selection overhead of only
0.114 hours, LeDAS maintains competitive computational
efficiency while delivering superior accuracy.

\cref{tab:imagenet1k} further demonstrates
the effectiveness of LeDAS across different architectures
and selection ratios. Notably, ResNet-50 matches its
full-dataset accuracy of 76.4\% with only 50\% of the training
samples selected per round, while ViT-Large surpasses
its full-dataset baseline even at a 20\% selection ratio.
At an 80\% selection ratio, LeDAS achieves 77.5\%,
84.2\%, and 85.4\% Top-1 accuracy on ResNet-50,
Swin-Base, and ViT-Large, respectively.
These results highlight the effectiveness of LeDAS
across diverse model architectures and data budgets.

\subsection{Effectiveness Across Diverse Tasks}    

\noindent\textbf{Object Detection.}
To evaluate the applicability of LeDAS beyond image
classification, we extend our framework to object detection
using YOLOv8 and RT-DETR on MS-COCO. Specifically, we
leverage sample-level detection losses to characterize
learning dynamics and guide data selection.
As shown in \cref{tab:coco}, LeDAS consistently
outperforms random selection across both detectors and all
selection ratios. Notably, with 90\% of the training data
selected per round, LeDAS achieves 40.8\% AP$_{50}$ and
28.0\% AP on YOLOv8, surpassing full-dataset training by
0.4 percentage points on both metrics. Similarly, RT-DETR
achieves 57.5\% AP$_{50}$ and 42.7\% AP, matching or
exceeding its full-dataset counterpart.
These results demonstrate the effectiveness and applicability
of LeDAS to dense prediction tasks.









\begin{table}[t]
    \centering
    \caption{Comparison of full-dataset supervised fine-tuning and LeDAS with 50\% of the training data on five benchmarks. Results are reported in Pass@1 (\%).}
    \label{tab:llm_sft}

    \resizebox{0.49\textwidth}{!}{
    \begin{tabular}{@{}llccccc@{}}
    \toprule[0.8pt]
    Model & Method & AIME24 & AMC23 & Olympiad & GPQA-D & MMLU-Pro \\
    \midrule[0.4pt]

    \multirow{2}{*}{Qwen3-4B}
    & Full Data
    & 36.67 & \textbf{72.50} & 36.87 & 54.15 & 63.69 \\
    & Ours (50\%)
    & \textbf{40.00} & \textbf{72.50} & \textbf{37.76}
    & \textbf{55.56} & \textbf{63.74} \\

    \midrule[0.4pt]

    \multirow{2}{*}{Qwen3.5-9B}
    & Full Data
    & \textbf{50.00} & 82.50 & 61.11 & 60.09 & 76.55 \\
    & Ours (50\%)
    & \textbf{50.00} & \textbf{85.00} & \textbf{61.57}
    & \textbf{62.63} & \textbf{77.12} \\

    \bottomrule[0.8pt]
    \end{tabular}}
    \vspace{-5pt}
\end{table}








\noindent\textbf{Semantic Segmentation.}
We further evaluate the applicability of LeDAS to semantic
segmentation using UPerNet and PSPNet on ADE20K.
As shown in \cref{tab:ade20k}, LeDAS
consistently outperforms random selection across both
architectures and all selection ratios. Notably, with
only 70\% of the training samples selected per round,
LeDAS achieves 33.52\% and 36.21\% mIoU on UPerNet
and PSPNet, surpassing their full-dataset baselines by
0.61 and 0.55 percentage points, respectively.
Furthermore, UPerNet achieves 33.85\% mIoU at an 80\%
selection ratio, while PSPNet reaches 36.35\% at 90\%.
These results demonstrate that LeDAS effectively extends
to pixel-level prediction tasks while maintaining
competitive performance under reduced data budgets.

\begin{table}[t]
    \centering
    \caption{Semantic segmentation performance on ADE20K
    using UPerNet and PSPNet under different selection
    ratios. Results are reported in mIoU (\%).}
    \label{tab:ade20k}

    \small
    \setlength{\tabcolsep}{4.5pt}
    \renewcommand{\arraystretch}{1.15}
    \resizebox{.8\columnwidth}{!}{
    \begin{tabular}{ll|ccc|c}
    \toprule[0.8pt]
    Model & Method & 70\% & 80\% & 90\% & Full Dataset \\
    \midrule[0.4pt]

    \multirow{2}{*}{UPerNet}
    & Random
    & 33.21 & 33.48 & 33.59
    & \multirow{2}{*}{32.91} \\
    & LeDAS
    & \textbf{33.52}
    & \textbf{33.85}
    & \textbf{33.71}
    & \\

    \midrule[0.4pt]

    \multirow{2}{*}{PSPNet}
    & Random
    & 35.64 & 35.77 & 36.12
    & \multirow{2}{*}{35.66} \\
    & LeDAS
    & \textbf{36.21}
    & \textbf{36.10}
    & \textbf{36.35}
    & \\
    \bottomrule[0.8pt]
    \end{tabular}}
    \vspace{-4pt}
\end{table}

\noindent\textbf{LLM Supervised Fine-Tuning.}
Beyond computer vision tasks, we further extend LeDAS to LLM supervised fine-tuning using Qwen3-4B and Qwen3.5-9B. To isolate the impact of data selection from differences in optimization duration, we use the same number of training steps for LeDAS and full-dataset fine-tuning, focusing on selection effectiveness rather than computational efficiency. As shown in \cref{tab:llm_sft}, LeDAS matches or outperforms full-dataset fine-tuning across all five benchmarks while selecting 50\% of the training samples per round. Specifically, Qwen3-4B achieves improvements of 3.33 and 0.89 percentage points on AIME24 and Olympiad, respectively, while matching the full-dataset performance on AMC23. For Qwen3.5-9B, LeDAS improves AMC23 performance by 2.50 percentage points and achieves comparable or better results on the other mathematical reasoning benchmarks. Moreover, LeDAS consistently improves performance on GPQA-Diamond and MMLU-Pro for both models, demonstrating its potential for cross-domain generalization. These results highlight the effectiveness of learning dynamics-aware data selection for LLM fine-tuning across different model scales.

\begin{table}[t]
    \centering
    \caption{Generalization performance of LeDAS on ImageNet-A/R/O/Hard. We report AUPR (\%) for ImageNet-O and accuracy (\%) for others using ResNet-50.}
    \label{tab:imagenet_generalization}

    \setlength{\tabcolsep}{4pt}
    \renewcommand{\arraystretch}{1.1}

    \resizebox{\columnwidth}{!}{
    \begin{tabular}{lcccc|c}
    \toprule
    Selection Ratio (\%) & 50 & 60 & 70 & 80 & Full Dataset \\
    \midrule

    ImageNet-A
    & 2.9$_{\downarrow 0.2}$
    & 3.0$_{\downarrow 0.1}$
    & 3.2$_{\uparrow 0.1}$
    & 3.7$_{\uparrow 0.6}$
    & 3.1 \\

    ImageNet-R
    & 37.4$_{\uparrow 1.2}$
    & 37.7$_{\uparrow 1.5}$
    & 38.1$_{\uparrow 1.9}$
    & 38.2$_{\uparrow 2.0}$
    & 36.2 \\

    ImageNet-O
    & 16.3$_{\uparrow 3.1}$
    & 16.4$_{\uparrow 3.2}$
    & 16.3$_{\uparrow 3.1}$
    & 16.4$_{\uparrow 3.2}$
    & 13.2 \\

    ImageNet-Hard
    & 14.9$_{\uparrow 0.2}$
    & 15.4$_{\uparrow 0.7}$
    & 16.0$_{\uparrow 1.3}$
    & 15.9$_{\uparrow 1.2}$
    & 14.7 \\

    \bottomrule
    \end{tabular}}
\end{table}

\subsection{Generalization on Challenging Benchmarks}

\noindent\textbf{Results on Challenging ImageNet Benchmarks.} To further assess the generalization and robustness of LeDAS, we evaluate it on four challenging benchmarks: ImageNet-A, ImageNet-R, ImageNet-O, and ImageNet-Hard. All models are trained on ImageNet-1K using either LeDAS-selected subsets or the full dataset, and the trained models are evaluated on these benchmarks. As shown in \cref{tab:imagenet_generalization}, LeDAS achieves consistent improvements across different selection ratios. Notably, even with only 50\% of the training data, LeDAS surpasses full-dataset training on ImageNet-R, ImageNet-O, and ImageNet-Hard by 1.2\%, 3.1\%, and 0.2\%, respectively. With a 70\% selection ratio, LeDAS further surpasses full-dataset training across all four benchmarks, achieving improvements of 0.1, 1.9, 3.1, and 1.3 percentage points on ImageNet-A, ImageNet-R, ImageNet-O, and ImageNet-Hard, respectively. At an 80\% selection ratio, these improvements increase to 0.6, 2.0, 3.2, and 1.2 percentage points, respectively. These results demonstrate that learning dynamics-aware selection can improve generalization to challenging evaluation distributions while substantially reducing the amount of training data.

\begin{table*}
    \begin{minipage}[]{0.6\textwidth}
        \centering
    \caption{Closed-set long-tailed classification accuracy (\%) on ImageNet-LT and Places-LT with an 80\% selection ratio.}
    \label{tab:long_tailed}

    \resizebox{0.99\textwidth}{!}{

    \small
    \setlength{\tabcolsep}{9pt}
    \renewcommand{\arraystretch}{1.2}

    \begin{tabular}{l|cccc|cccc}
    \toprule[0.8pt]

    \multirow{2}{*}{Method}
    & \multicolumn{4}{c|}{ImageNet-LT}
    & \multicolumn{4}{c}{Places-LT} \\

    \cline{2-9}

    & Many & Med. & Few & Acc.
    & Many & Med. & Few & Acc. \\

    \midrule[0.4pt]

    Full Dataset
    & \textbf{71.83} & 54.50 & 33.49 & 57.85
    & \textbf{41.26} & 34.21 & 19.74 & 34.99 \\

    LeDAS
    & 71.62 & \textbf{55.23} & \textbf{33.81} & \textbf{58.12}
    & 40.60 & \textbf{35.01} & \textbf{20.95} & \textbf{35.55} \\

    \bottomrule[0.8pt]
    \end{tabular}}
    \end{minipage}
    \hspace{12pt}
    \begin{minipage}[]{0.3\textwidth}
     
        \centering
    \caption{Robustness to noisy and corrupted data on
    Tiny-ImageNet using ResNet-50.}
    \label{tab:noisy_corrupted}
    \small
    \resizebox{0.99\textwidth}{!}{
    \begin{tabular}{c|cc|cc}
        \toprule[0.8pt]
        \multirow{2}{*}{Method}
        & \multicolumn{2}{c|}{Noisy Dataset}
        & \multicolumn{2}{c}{Corrupted Dataset} \\
        \cmidrule(lr){2-3}
        \cmidrule(lr){4-5}
        & 20\% & 30\% & 20\% & 30\% \\
        \midrule[0.4pt]
        Random
        & 17.8 & 23.9 & 20.0 & 25.9 \\

        Random*
        & 32.5 & 36.1 & 36.5 & 38.3 \\

        CLIP-Sel
        & 26.1 & 33.1 & 26.1 & 32.1 \\

        LeDAS
        & \textbf{39.9} & \textbf{45.6}
        & \textbf{42.6} & \textbf{47.1} \\
        \bottomrule[0.8pt]
        \end{tabular}}
    \end{minipage}
\end{table*}

\noindent\textbf{Results on Long-Tailed Datasets.}
We evaluate the effectiveness of LeDAS under imbalanced class distributions by conducting closed-set long-tailed classification experiments on ImageNet-LT and Places-LT using ViT-B with an 80\% selection ratio. As reported in \cref{tab:long_tailed}, LeDAS improves overall accuracy on both datasets while utilizing fewer training samples. On ImageNet-LT, it boosts the overall accuracy from 57.85\% to 58.12\%, yielding gains of 0.73 and 0.32 percentage points on medium- and few-shot classes, respectively. Similarly, on Places-LT, it improves overall accuracy from 34.99\% to 35.55\%, accompanied by gains of 0.80 and 1.21 percentage points on medium- and few-shot classes. Despite a slight drop in many-shot classes, these results demonstrate that LeDAS better preserves recognition performance on underrepresented classes while improving overall accuracy under a reduced data budget.

\noindent\textbf{Results on Noisy and Corrupted Datasets.}
The modular design of LeDAS supports the integration of task-specific signals beyond conventional clean-data settings. Specifically, we incorporate a cross-modal semantic alignment signal~\cite{yang2024clip} to capture visual-textual consistency, providing complementary supervision for sample selection under noisy and corrupted conditions. We evaluate LeDAS on Tiny-ImageNet with both noise and corruption ratios set to 20\%. As shown in Table~\ref{tab:noisy_corrupted}, LeDAS consistently outperforms existing baselines across both settings and selection ratios. On the noisy dataset, LeDAS achieves 39.9\% and 45.6\% accuracy at 20\% and 30\% selection ratios, respectively, surpassing existing SOTA baselines by 7.4 and 9.5 percentage points. On the corrupted dataset, it achieves 42.6\% and 47.1\% accuracy, with improvements of 6.1 and 8.8 percentage points, respectively. These results demonstrate the extensibility and robustness of LeDAS to diverse data quality conditions without modifying its core selection framework.

\begin{table}[t]
    \centering
    \caption{Ablation study of the three learning dynamics
    metrics on Tiny-ImageNet using ResNet-50.
    $H$, $P$, and $D$ denote hardness, progress, and
    deterioration, respectively.}
    \label{tab:ablation_hpd}

    \resizebox{0.35\textwidth}{!}{

    \small
    \setlength{\tabcolsep}{10pt}
    \renewcommand{\arraystretch}{1.15}

    \begin{tabular}{ccc|ccc}
        \toprule
        $H$ & $P$ & $D$
        & 30\% & 50\% & 70\% \\
        \midrule

        & & & 41.1 & 42.8 & 43.1 \\

        $\checkmark$ & &
        & 45.8 & 48.0 & 49.5 \\

        & $\checkmark$ &
        & 45.7 & 48.2 & 49.7 \\

        & & $\checkmark$
        & 45.4 & 49.0 & 49.5 \\

        $\checkmark$ & $\checkmark$ &
        & 45.8 & 48.9 & 50.0 \\

        $\checkmark$ & & $\checkmark$
        & 45.7 & 47.9 & 50.0 \\

        & $\checkmark$ & $\checkmark$
        & 45.4 & 48.5 & 50.1 \\

        \midrule
        $\checkmark$ & $\checkmark$ & $\checkmark$
        & \textbf{45.9} & \textbf{49.1} & \textbf{50.2} \\
        \bottomrule
    \end{tabular}}
\end{table}

\subsection{Ablation Studies}

\noindent\textbf{Effect of Learning Dynamics Metrics.}
An ablation study on Tiny-ImageNet using ResNet-50 examines the contributions of individual learning dynamics metrics across all seven combinations of hardness, progress, and deterioration. As reported in \cref{tab:ablation_hpd}, each metric individually improves classification accuracy over the baseline without these metrics, demonstrating the effectiveness of different learning signals for data selection. Notably, combining all three metrics achieves the highest accuracy across all selection ratios, reaching 45.9\%, 49.1\%, and 50.2\% at 30\%, 50\%, and 70\% selection ratios, respectively. Compared with the strongest two-metric combinations, the complete formulation further improves accuracy by 0.1, 0.2, and 0.1 percentage points, respectively. These results highlight the benefits of jointly modeling sample difficulty and directional learning dynamics for effective data selection.

\begin{figure}[t]
    \centering
    \includegraphics[width=\linewidth]{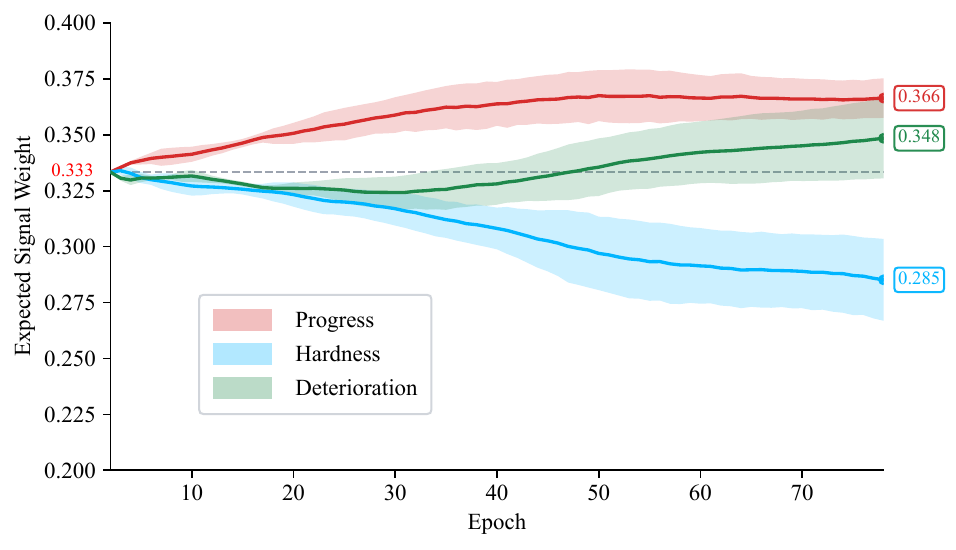}
    \caption{Evolution of adaptive metric fusion weights
    on Tiny-ImageNet with a 30\% selection ratio.
    Solid lines denote the mean expected weights, and the dashed line denotes uniform weighting.}
    \label{fig:signal_weight}
    \vspace{-5pt}
\end{figure}

\begin{table}[t]
    \centering
    \caption{Ablation study of adaptive metric fusion and
    staleness compensation on Tiny-ImageNet using ResNet-50.}
    \label{tab:ablation_components}

    \small
    \setlength{\tabcolsep}{9pt}
    \renewcommand{\arraystretch}{1.15}
    \resizebox{0.43\textwidth}{!}{
    \begin{tabular}{lccc}
        \toprule
        Method & 30\% & 50\% & 70\% \\
        \midrule
        w/o Adaptive Metric Fusion
        & 45.5 & 48.7 & 49.9 \\
        w/o Staleness Compensation
        & 45.7 & 48.1 & 50.0 \\
        \midrule
        LeDAS
        & \textbf{45.9}
        & \textbf{49.1}
        & \textbf{50.2} \\
        \bottomrule
    \end{tabular}}
    \vspace{-5pt}
\end{table}

\noindent\textbf{Effect of Adaptive Components.}
The roles of Adaptive Metric Fusion and Staleness Compensation are further examined through ablation experiments on Tiny-ImageNet using ResNet-50. As shown in \cref{tab:ablation_components}, removing either component consistently degrades performance across all selection ratios. Specifically, removing Adaptive Metric Fusion reduces accuracy by 0.4, 0.4, and 0.3 percentage points at 30\%, 50\%, and 70\% selection ratios, respectively. Similarly, removing Staleness Compensation leads to accuracy drops of 0.2, 1.0, and 0.2 percentage points.

We further visualize the evolution of learned metric weights in \cref{fig:signal_weight}. Starting from approximately uniform weighting, the bandit policy gradually increases the emphasis on progress and deterioration while reducing the contribution of hardness. This trend suggests that the relative importance of different learning signals evolves throughout training, supporting the motivation for adaptive metric fusion rather than fixed weighting. Together, these results demonstrate the effectiveness of both adaptive fusion and staleness compensation in improving data selection.

\section{Conclusion}
\label{sec:conclusion}

In this paper, we propose LeDAS, an end-to-end learning dynamics-aware data selection framework. By jointly characterizing sample hardness, progress, and deterioration, LeDAS captures evolving sample importance through Adaptive Metric Fusion and bounded Staleness Compensation. Extensive experiments across diverse vision tasks and LLM supervised fine-tuning demonstrate that LeDAS consistently matches or surpasses existing methods while using substantially less training data. These results highlight the effectiveness and broad applicability of learning dynamics-aware data selection for data-efficient model training.

\clearpage
\section*{LLM Usage Statement}
DeepSeek was used to assist with language polishing, improving academic phrasing, and addressing LaTeX syntax and formatting issues. All AI-assisted suggestions were reviewed and revised by the authors, who take full responsibility for the final manuscript.
{
    \small
    \bibliographystyle{ieeenat_fullname}
    \bibliography{main}
}


\end{document}